\documentclass[10pt,twocolumn,letterpaper]{article}

\usepackage{cvpr}
\usepackage{times}
\usepackage{epsfig}
\usepackage{graphicx}
\usepackage{amsmath}
\usepackage{amssymb}
\usepackage{multirow}
\usepackage{subfigure}
\usepackage{chngpage}
\usepackage{xcolor,colortbl}
\usepackage{bbold}
\usepackage{dsfont}
\usepackage{booktabs}

\usepackage[pagebackref=true,breaklinks=true,letterpaper=true,colorlinks,bookmarks=false]{hyperref}

\cvprfinalcopy 

\def\cvprPaperID{2017} 

\newcommand{\myparagraph}[1]{{\vspace{0.5em} \noindent \bf #1}}

\usepackage{caption}
\begin{document}

\title{Associatively Segmenting Instances and Semantics in Point Clouds}

\author{Xinlong Wang$^{1}$ \quad Shu Liu$^2$ \quad Xiaoyong Shen$^2$ \quad Chunhua Shen$^1$ \quad Jiaya Jia$^{2,3}$ \\
\and 
$^1$The University of Adelaide\\
\and
$^2$Youtu Lab, Tencent\\
\and
$^3$The Chinese University of Hong Kong\\
{\tt\small  \{wangxinlon, liushuhust, goodshenxy\}@gmail.com}\\
{\tt\small chunhua.shen@adelaide.edu.au~~ leojia@cse.cuhk.edu.hk}\\
}

\maketitle

\begin{abstract}
  A 3D point cloud describes the real scene precisely and intuitively.
  To date how to segment diversified elements in such an informative 3D scene is rarely discussed. 
  In this paper, we first introduce a simple and flexible framework to segment instances and semantics in point clouds simultaneously.
  Then, we propose two approaches which make the two tasks take advantage of each other, leading to a win-win situation.
  Specifically, we make instance segmentation benefit from semantic segmentation through learning semantic-aware point-level instance embedding.
  Meanwhile, semantic features of the points belonging to the same instance are fused together to make more accurate per-point semantic predictions. 
  Our method largely outperforms the state-of-the-art method in 3D instance segmentation along with a significant improvement in 3D semantic segmentation.
  Code has been made available at: 
  \href{https://github.com/WXinlong/ASIS}{https://github.com/WXinlong/ASIS}.

\end{abstract}


\section{Introduction}

Both instance segmentation and semantic segmentation aim to detect specific informative region represented by sets of smallest units in the scenes.
For example, a point cloud can be parsed into groups of points, where each group corresponds to a class of stuff or an individual instance.
The two tasks are related and both have wide applications in real scenarios, \eg, autonomous driving and augmented reality.
Though great progress has been made in recent years~\cite{he2017mask,de2017semantic,long2015fully,tchapmi2017segcloud,Landrieu_2018_CVPR} for each single task, no prior method tackles these two tasks associatively.

In fact, instance segmentation and semantic segmentation conflict with each other in some respects.
The former one distinguishes different instances of the same class clearly, while the latter one wants them to have the same label.
However, the two tasks could cooperate with each other through seeking common grounds.
Semantic segmentation distinguishes points of different classes, which is also one of the purposes of instance segmentation, as 
{\it points of different classes must belong to different instances.}
Furthermore, instance segmentation assigns the same label to points belonging to the same instance, which is also consistent with semantic segmentation, as 
{\it points of the same instance must belong to the same category.} 
This observation makes one wonder how the two tasks could be associated together to lead to a win-win solution?

\begin{figure}[!tb]
\includegraphics[width=0.48\textwidth,height=0.24\textwidth]{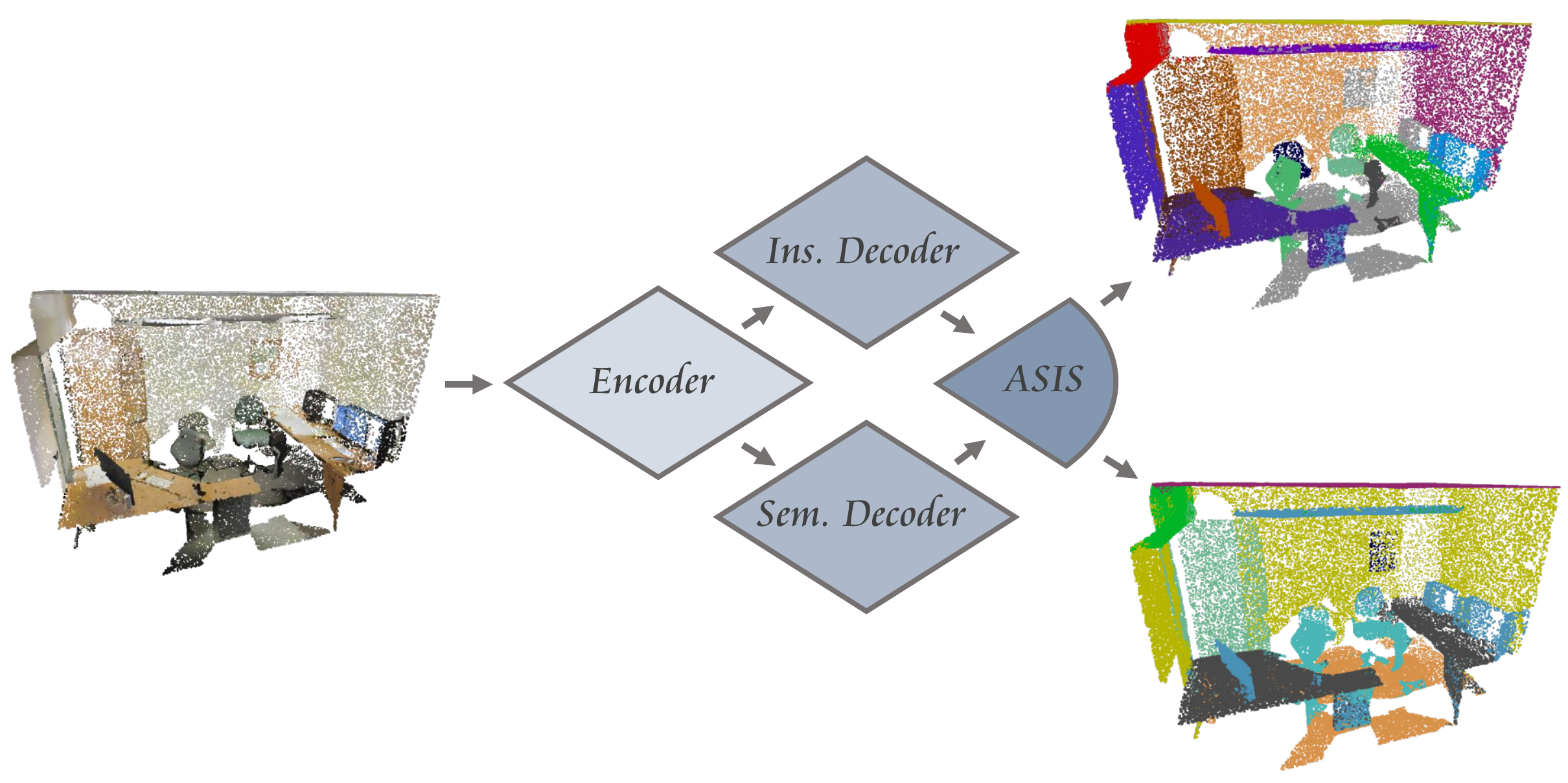}
\caption{Instance segmentation and semantic segmentation results using ASIS. Our method takes raw point clouds as inputs and outputs instance labels and semantic labels for each point.}
\label{fig:intro}
\end{figure}

There may be two  straightforward approaches. 
The first one is that, given the semantic labels, we could run instance segmentation independently on every semantic class to better distinguish individual instances. Thus,  different class instances are separated simply but naively.

However, the instance segmentation would greatly depend on performance of semantic segmentation as incorrect semantic predictions would inevitably result in incorrect instance predictions. 
Otherwise, given the instance labels, one could classify each instance and assign the predicted class label to each point of this instance. Thus, the problem is transformed to an easier instance recognition problem. However, inaccurate instance predictions would deeply confuse the  downstream object classifiers.
Both these two approaches are in step-wise paradigm, which can be sub-optimal and inefficient.
In this work, we integrate  the two tasks altogether into an end-to-end parallel training framework, which shares the same benefits in a soft and learnable fashion.

We first introduce a simple baseline to segment instances and semantics simultaneously. 
It is similar to the method in~\cite{de2017semantic} for 2D images, but we tailor it for 3D point cloud.
The network of the baseline has two parallel branches: one for per-point semantic predictions; the other outputs point-level instance embeddings, where the embeddings of points belonging to the same instance stay close while those of different instances are apart.
Our baseline method can already achieve better performance than the recent state-of-the-art method, SGPN~\cite{sgpn}, as well as faster training and inference.
Based on this flexible baseline, a novel technique is further proposed to associate instance segmentation and semantic segmentation closely together, termed ASIS (Associatively Segmenting Instances and Semantics).

\begin{figure}[!tb]
\includegraphics[width=0.48\textwidth,height=0.24\textwidth]{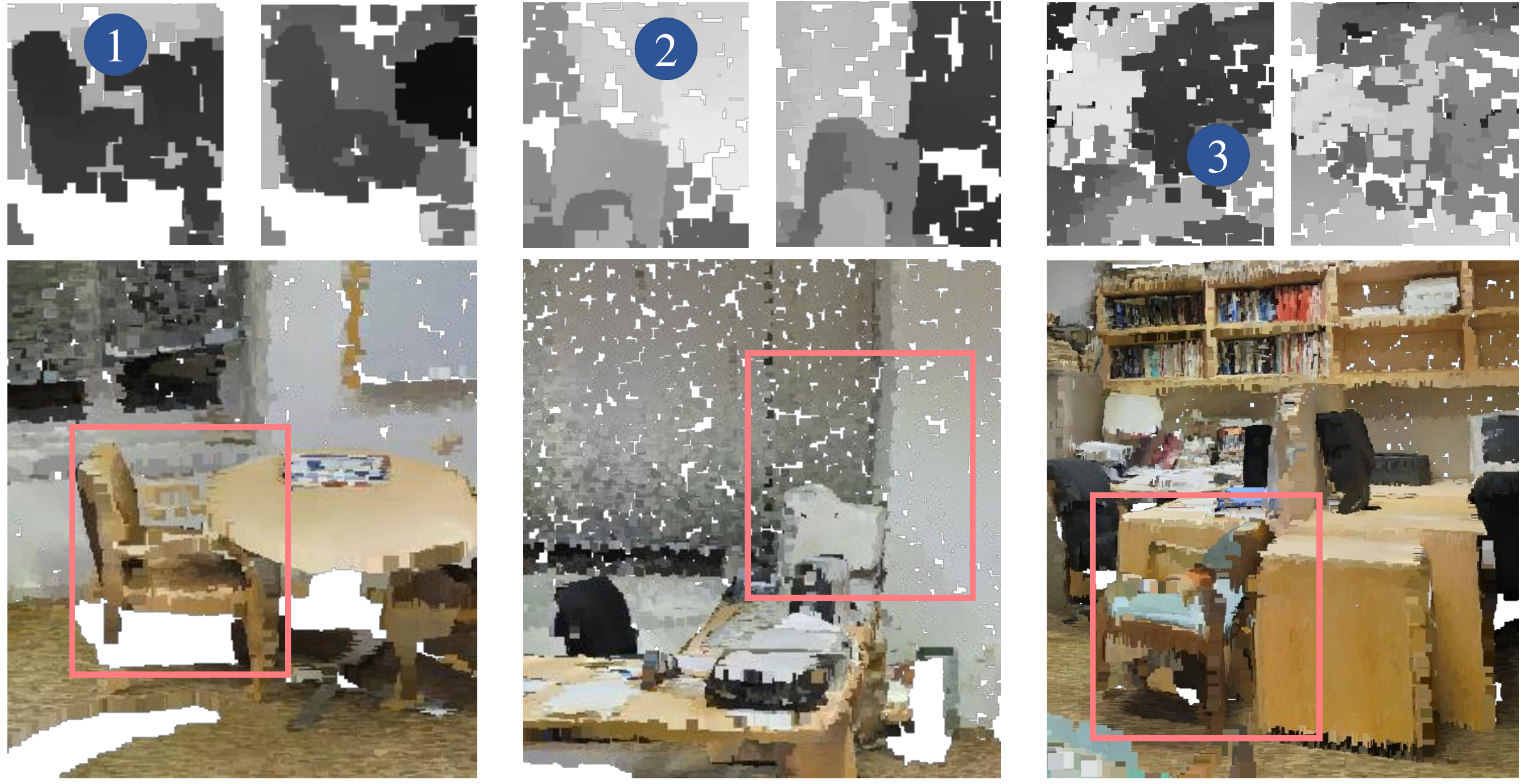}
\caption{1D embeddings of learned point-level instance embeddings. t-SNE~\cite{maaten2008visualizing} technique is used to visualize the learned instance embeddings for the points on {\em S3DIS test} data. Three close-up  pairs are shown. Of each pair the left patch is from our baseline method, while the right one is from ASIS. Differences in color shade represent distances in instance embedding space.}
\label{fig:tsne}
\end{figure}

With the proposed ASIS method, we are able to learn semantic-aware instance embeddings, where the embeddings of points belonging to different semantic classes are further separated automatically through feature fusion.
As shown in Figure~\ref{fig:tsne}, the boundaries between different class points are clearer (chair \& table, window \& wall). 
Moreover, the semantic features of points belonging to the same instance are  exploited 
and fused together to make more accurate per-point semantic predictions. 
The intuition behind it is that during semantic segmentation {\it a point being assigned to one of the categories is because the instance containing that point belongs to that category}.
Thus, the two tasks can take advantage of each other to further boost their performance.
Our method is demonstrated to be effective and general on different backbone networks, \eg, the PointNet~\cite{Qi_2017_CVPR} and hierarchical architecture PointNet++~\cite{qi2017pointnet++}.
The method can also be used to tackle the panoptic segmentation~\cite{kirillov2018panoptic} task, which unifies the semantic and instance segmentation. 
To summarize, our main contributions are as follows.
\begin{itemize}
\setlength{\itemsep}{0pt}
\setlength{\parskip}{0pt}
\setlength{\parsep}{0pt}
  \item We 
  propose a fast and efficient simple baseline for simultaneous instance segmentation and semantic segmentation on 3D point clouds. 
  \item We propose a new framework, termed ASIS, to associate instance segmentation and semantic segmentation closely together. Specifically, two types of partnerships are proposed---semantics  awareness for instance segmentation and instance fusion for semantic segmentation---to make these two tasks cooperate with each other.
  \item With the proposed ASIS, the model containing semantics-aware instance segmentation and instance-fused semantic segmentation are trained end-to-end, which outperforms  state-of-the-art 3D instance segmentation methods on the S3DIS dataset~\cite{s3dis} along with a significant improvement on the 3D semantic segmentation task. 
  Furthermore, our experiments on the ShapeNet dataset~\cite{shapenet} show that ASIS is also beneficial for the task of part segmentation.
\end{itemize}

\section{Related Work}

\myparagraph{Instance Segmentation.}
2D instance segmentation has attracted much research attention recently, leading to various top-performing methods. 
Inspired by the effectiveness of region-based CNN (R-CNN)~\cite{Girshick_2014_CVPR} in object detection problem, \cite{DeepMask, dai2016instance1} learn to segment instances by proposing segment candidates. The mask proposals are further classified to obtain the final instance masks. 
Dai \etal~\cite{dai2016instance2} predict segment proposals based on bounding box proposals. 
He \etal~\cite{he2017mask} propose the simpler and flexible Mask R-CNN which predicts masks and class labels simultaneously.
Different from the top-down detector-based approaches above, bottom-up methods learn to associate per-pixel predictions to object instances.
Newell \etal~\cite{associativeembedding} group pixels into instances using the learned associative embedding. 
Brabandere \etal~\cite{de2017semantic} propose a discriminative loss function which enables to learn pixel-level instance embedding efficiently. 
Liu \etal~\cite{SGN17} decompose the instance segmentation problem into a sequence of sub-grouping problems.
However, 3D instance segmentation is rarely researched. 
Wang \etal~\cite{sgpn} learn the similarity matrix of a point cloud to get instance proposals. 
In this work, we introduce a simple and flexible method that learns effective point-level instance embedding with the help of semantic features in 3D point clouds.

\myparagraph{Semantic Segmentation.}
With the recent development of convolutional neural networks (CNNs)~\cite{krizhevsky2012imagenet,simonyan2014very},  tremendous  progress has been made in semantic segmentation. 
Approaches~\cite{lin2017refinenet,chen2017rethinking,Lin2017Semantic} based on fully convolutional networks (FCN)~\cite{long2015fully} dominate the semantic segmentation on 2D images. 
As for 3D segmentation, Huang \etal~\cite{3dfcnn} propose 3D-FCNN which predict coarse voxel-level semantic label.
PointNet~\cite{Qi_2017_CVPR} and following works~\cite{3dsemseg_ICCVW17,Ye_2018_ECCV} use multilayer perceptron (MLP) to produce fine-grained point-level segmentation.
Very recently, Landrieu \etal~\cite{Landrieu_2018_CVPR} introduce superpoint graph (SPG) to segment large-scale point clouds.
In fact, few of previous works segment semantics taking advantages of the instance embedding, either in 2D images or 3D point clouds.

\myparagraph{Deep Learning on Point Clouds.}
To take advantage of the  strong representation capability of classic CNNs, a 3D point cloud is first projected into multiview rendering images in~\cite{su2015multi,shideeppano,qi2016volumetric,guerry2017snapnet}, on which the well-designed CNNs for 2D images can be applied.
But part of contextual information in point cloud is left behind during the projection process.
Another popular representation for point cloud data is voxelized volumes. The works of
\cite{wu20153d,maturana2015voxnet,jingicpr,riegler2017octnet} convert point cloud data into regular volumetric occupancy grids, then train 3D CNNs or the varieties to perform voxel-level predictions. 
A drawback of volumetric representations is being both computationally and memory intensive, due to the sparsity of point clouds and the heavy computation of 3D convolutions. Therefore
those methods are limited to deal with large-scale 3D scenes.
To process raw point cloud directly, PointNet~\cite{Qi_2017_CVPR} is proposed to yield point-level predictions, achieving strong performance on 3D classification and segmentation tasks.
The following works PointNet++~\cite{qi2017pointnet++}, RSNet~\cite{huang2018recurrent}, DGCNN~\cite{dgcnn} and PointCNN~\cite{li2018pointcnn} further focus on exploring the local context and hierarchical learning architectures.
In this work, we build a novel framework to associatively segment instances and semantics in point clouds, and demonstrate that it is effective and general on different backbone networks.

\begin{figure*}[htbp]
\centering
\subfigure[]{
\includegraphics[width=0.54\textwidth]{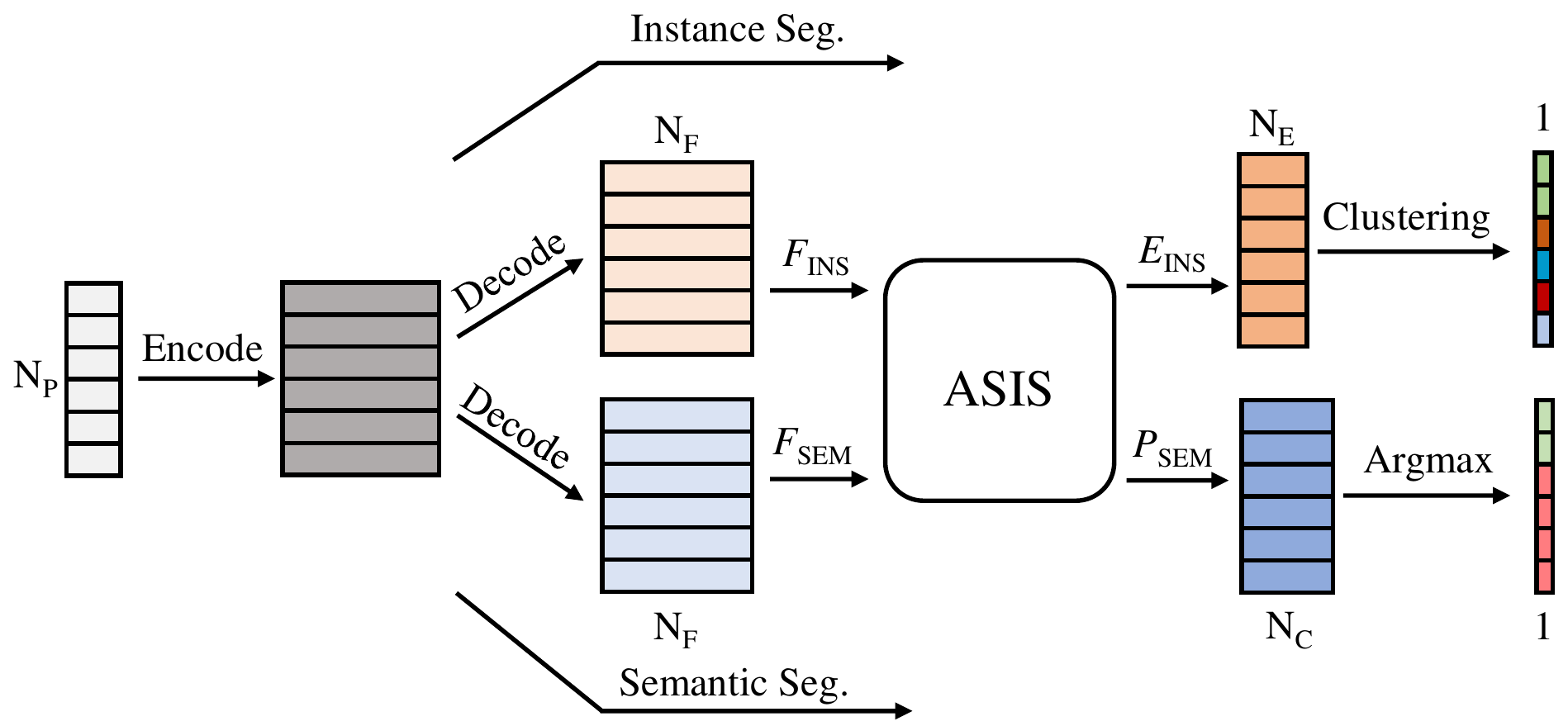}
\label{fig:framework1}
}
\hspace{.3in}
\subfigure[]{
\includegraphics[width=0.38\textwidth]{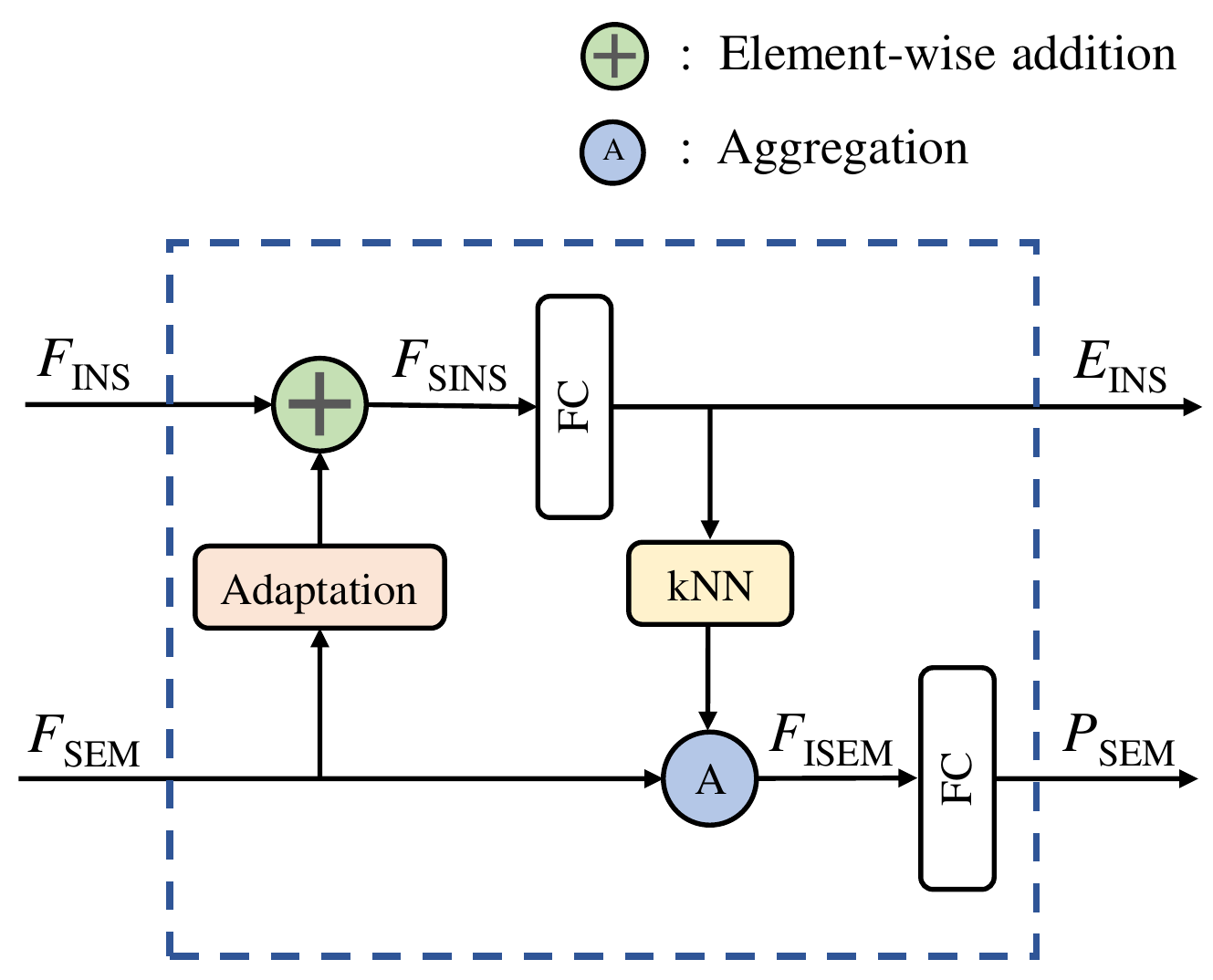}
\label{fig:framework2}
}
\caption{Illustration of our method for point cloud instance segmentation and semantic segmentation. (a) Full pipeline of the system.   (b) Illustration of the ASIS module.}
\label{fig:framework}
\end{figure*}

\section{Our Method}

\subsection{A Simple Baseline}
\label{subsec:baseline}

Here we introduce a simple yet effective framework.
It is composed of a shared encoder and two parallel decoders.
One of the decoders is for point-level semantic predictions, while the other one aims to handle the instance segmentation problem. 
Specifically, a point cloud of size $N_p$ is first extracted and encoded  into a feature matrix through the feature encoder (\eg, stacked PointNet layers). 
This shared feature matrix refers to the concatenation of local features and global features in PointNet architecture, or the output of the last set abstraction module for the PointNet++ architecture.
The two parallel branches then fetch the feature matrix and proceed with their following predictions separately.
The semantic segmentation branch decodes the shared feature matrix into $N_P \times N_{F}$ shaped  semantic feature matrix $F_{\rm SEM}$, and then outputs the semantic predictions $P_{\rm SEM}$ with shape of $N_P \times N_C$ where $N_C$ is the number of semantic categories. 
The instance segmentation branch has the same architecture except the last output layer.
The $N_P \times N_{F}$ instance feature matrix $F_{\rm INS}$ is used to predict per-point instance embedding $E_{\rm INS}$ with shape of $N_P \times N_{E}$ where $N_{E}$ is the dimension of the embedding.
The embeddings of a point cloud represent the the instance relationship between points in it: the points belonging to the same instance are close to each other in embedding space, while those points of different instances are apart. 

At training time, the semantic segmentation branch is supervised by the classical cross entropy loss. 
As for the instance segmentation, the discriminative loss function for 2D image in~\cite{de2017semantic} is adopted to supervise the instance embedding learning.
We modify it and make it suitable for point clouds.
The loss used in~\cite{de2017semantic} is class-specific: the instance embeddings of different semantic class are learned separately, which means the semantic class should be given first.
This step-wise paradigm is highly dependent on the quality of semantic prediction, as incorrect semantic prediction would inevitably result in incorrect instance recognition.
Thus, we adopt the class-agnostic instance embedding learning strategy, where embeddings are in charge of distinguishing different instances and are blind to their categories. The loss function is formulated as follows:
\begin{equation}
L =  L_{var} + L_{dist} + \alpha \cdot L_{reg},
\end{equation}
where $L_{var}$ aims to pull embeddings towards the mean embedding of the instance, \ie the instance center, 
$L_{dist}$ make instances repel each other, and $L_{reg}$ is a regularization term to keep the embedding values bounded. $\alpha$ is set to $0.001$ in our experiments. Specifically, each term can be written as follows:
\begin{equation}
	L_{var} = \frac{1}{I} \sum_{i=1}^{I} \frac{1}{N_i} \sum_{j=1}^{N_i} \left[ {\lVert \mu_i - e_j \rVert}_1 - \delta_{\textrm{v}} \right]_{+}^2,
\label{eq:l_var}
\end{equation}
\begin{equation}
L_{dist} = \frac{1}{I (I-1)} \mathop{\sum_{i_A = 1}^{I} \sum_{i_B = 1}^{I}}_{i_A \neq i_B} \left[ 2 \delta_{\textrm{d}} - {\lVert \mu_{i_A} - \mu_{i_B} \rVert}_1 \right]_{+}^2,
\end{equation}
\begin{equation}
L_{reg} = \frac{1}{I} \sum_{i=1}^{I} {\lVert \mu_{i} \rVert}_1,
\end{equation}
where $I$ is the number of ground-truth instances; $N_i$ is the number of points in instance $i$;
$\mu_i$ is the mean embedding of instance $i$;
${\lVert \cdot \rVert}_1$ is the $\ell_1$ distance;
$e_j$ is an embedding of a point;
$\delta_{\textrm{v}}$ and $\delta_{\textrm{d}}$ are margins;
$[x]_{+} = \max (0, x)$ means the hinge.

During the test, 
final instance labels are obtained using mean-shift clustering~\cite{comaniciu2002mean} on instance embeddings.
We assign the mode of the semantic labels of the points within the same instance as its final category. The pipeline is illustrated in Figure~\ref{fig:framework1}.

\subsection{Mutual Aid}
As depicted in Figure~\ref{fig:framework2}, benefiting from the simple and flexible framework described above, we are able to build upon it the novel ASIS module and achieve semantic-aware instance segmentation and instance-fused semantic segmentation. 

\myparagraph{Semantic-aware Instance Segmentation.}
Semantic features of a point cloud construct a new and high-level feature space, where points are naturally positioned according to their categories.
In that space, points of the same semantic class lie close together while different classes are separated.
We abstract the semantic awareness (SA) from semantic features and integrate it into the instance features, producing semantic-aware instance features. 
Firstly, the semantic feature matrix $F_{\rm SEM}$ is adapted to instance feature space as $F_{\rm SEM}'$ through a point independent fully connected layer (FC) with batch normalization and ReLU activation function.
$F_{\rm SEM}'$ has the same shape with $F_{\rm SEM}$.
Then, We add the adapted semantic feature matrix $F_{\rm SEM}'$ to instance feature matrix $F_{\rm INS}$ element-wise, producing semantic-aware instance feature matrix $F_{\rm SINS}$ .
The procedure can be formulated as:
\begin{equation}
    F_{\rm SINS} = F_{\rm INS} + {FC}(F_{\rm SEM}).
\end{equation}
In this soft and learnable way, points belonging to different category instances are further repelled in instance feature space, whereas same category instances are rarely affected.
The feature matrix $F_{\rm SINS}$ is used to generate final instance embeddings.

\myparagraph{Instance-fused Semantic Segmentation.}
Given the instance embeddings, we use $K$ nearest neighbor ($k$NN) search to find a fixed number of neighboring points for each point (including itself) in instance embedding space. 
To make sure the $K$ sampled points belonging to the same instance, 
we filter the outliers according to the margin $\delta_{v}$ used in Equation~\ref{eq:l_var}.
As described in Section~\ref{subsec:baseline}, the hinged loss term $L_{var}$ supervises the instance embedding learning through drawing each point embedding close to the the mean embedding within a distance of $\delta_{v}$.
The output of the kNN search is an index matrix with shape of $N_P \times K$.
According to the index matrix, the semantic features ($F_{\rm SEM}$) of those points are grouped to a $N_P \times K \times N_{F}$ shaped feature tensor, which is groups of semantic feature matrix where each group corresponds to a local region in instance embedding space neighboring its centroid point.
Inspired by the effectiveness of channel-wise max aggregation in~\cite{Qi_2017_CVPR,dgcnn,Ye_2018_ECCV}, semantic features of each group are fused together through a channel-wise max aggregation operation, as the refined semantic feature of the centroid point.
The instance fusion (IF) can be formulated as below. For the $ N_P \times N_{F}$ shaped semantic feature matrix $F_{\rm SEM} = \{ x_1, ..., x_{N_P} \} \subseteq \mathbb{R}^{N_F} $, instance-fused semantic features are calculated as:
\begin{equation}
    x_{i}^{\prime} = {\rm Max}(x_{i1}, x_{i2}, ..., x_{ik}),
\end{equation}
where $\{x_{i1}, ..., x_{ik} \}$ represent the semantic features of $K$ neighboring points centered point $i$ in instance embedding space, and $\rm Max$ is an element-wise maximum operator which takes $K$ vectors as input and output a new vector. 
After the instance fusion, the output is a $N_P \times N_{F}$ feature matrix $F_{\rm ISEM}$, the final semantic features to be fed into the last semantic classifier.

\section{Experiments}

\subsection{Experiment Settings}
\myparagraph{Datasets.}
We carry out experiments on two public datasets: Stanford 3D Indoor Semantics Dataset (S3DIS)~\cite{s3dis} and ShapeNet~\cite{shapenet}. 
S3DIS contains 3D scans from Matterport Scanners in $6$ areas that in total have $272$ rooms.
Each point in the scene point cloud is associated with an instance label and one of the semantic labels from $13$ categories.
Besides the large real scene benchmark S3DIS, we also evaluate our methods on the ShapeNet part dataset.
This dataset contains $16,881$ 3D shapes from $16$ categories. Each point sampled from the shapes is assigned with one of the $50$ different parts. The instance annotations from~\cite{sgpn} are used as the instance ground-truth labels.

\myparagraph{Evaluation Metrics.}
Our experiments involved S3DIS are conducted following the same k-fold cross validation with micro-averaging as in~\cite{Qi_2017_CVPR}. 
We also report the performance on the fifth fold following~\cite{tchapmi2017segcloud}, as Area 5 is not present in other folds. 
For evaluation of semantic segmentation, overall accuracy (oAcc), mean accuracy (mAcc) and mean IoU (mIoU) across all the categories are calculated along with the detailed scores of per class IoU.
As for instance segmentation, (weighted) coverage (Cov, WCov)~\cite{ren17recattend, SGN17,zhuo2017indoor} are adopted.
Cov is the average instance-wise IoU of prediction matched with ground-truth. 
The score is further weighted by the size of the ground-truth instances to get WCov.
For ground-truth regions $\mathcal{G}$ and predicted regions $\mathcal{O}$, these values are defined as 
\begin{equation}
    \textrm{Cov}(\mathcal{G}, \mathcal{O}) = \sum_{i=1}^{|\mathcal{G}|}\frac{1}{|\mathcal{G}|}  \max_{j} \textrm{IoU}(r_{i}^{G}, r_{j}^{O}),
\end{equation}

\begin{equation}
    \textrm{WCov}(\mathcal{G}, \mathcal{O}) = \sum_{i=1}^{|\mathcal{G}|}w_{i}  \max_{j} \textrm{IoU}(r_{i}^{G}, r_{j}^{O}),
\end{equation}

\begin{equation}
     w_{i} = \frac{|r_{i}^{G}|}{\sum_k |r_{k}^{G}|},
\end{equation}
where $|r_i^G|$ is the number of points in ground-truth region $i$.
Besides, the classical metrics mean precision (mPrec) and mean recall (mRec) with IoU threshold $0.5$ are also reported. 

\myparagraph{Training and Inference Details.}
For the S3DIS dataset, each point is represented by a $9$-dim feature vector (XYZ, RGB and normalized coordinates as to the room).
During training, we follow the procedure in~\cite{Qi_2017_CVPR} and split the rooms into $1m \times 1m$ overlapped blocks on the ground plane, each containing $4096$ points.
For the instance segmentation branch, we train the network with ${\sigma}_v = 0.5$, ${\sigma}_d = 1.5$, and $5$ output embedding dimensions. 
For the $k$NN search in instance fusion, K is set to $30$.
We train the network for 50 epochs and 100 epochs for PointNet and PointNet++ respectively, with batch size $24$, base learning rate set to $0.001$ and divided by 2 every $300k$ iterations. 
The Adam solver is adopted to optimize the network on a single GPU.
Momentum is set to 0.9.
At test time, bandwidth is set to $0.6$ for mean-shift clustering.
BlockMerging algorithm~\cite{sgpn} is used to merge instances from different blocks.
For ShapeNet dataset, each shape is represented by a point cloud with $2048$ points, as in~\cite{Qi_2017_CVPR}. Each point is represented by a $3$-dim vector ($XYZ$).

\begin{table}[!ht]
\begin{center}
\small 
\setlength{\tabcolsep}{3.8pt}
\begin{tabular}{c|c|c|c|c|c}
\hline 
\hline
 Backbone & Method    & mCov    & mWCov    &  mPrec & mRec \\
\hline
\hline
\multicolumn{6}{c}{Test on Area 5} \\
\hline
\multirow{3}{*}{PN} 
  & SGPN ~\cite{sgpn} &  32.7  & 35.5  &  36.0  & 28.7   \\
  & ASIS (\textit{vanilla}) & 38.0 & 40.6 & 42.3 & 34.9 \\
  & ASIS & {\bf 40.4}  & {\bf43.3} & {\bf 44.5} & {\bf 37.4} \\
 \hline
  \multirow{2}{*}{PN++}  
  & ASIS (\textit{vanilla}) & 42.6 & 45.7 & 53.4 & 40.6 \\
  & ASIS & {\bf 44.6}  & {\bf 47.8} & {\bf 55.3} & {\bf 42.4} \\
  
\hline 
\hline
\multicolumn{6}{c}{Test on 6-fold CV} \\
\hline
 \multirow{3}{*}{PN} 
 & SGPN~\cite{sgpn} & 37.9   & 40.8  & 38.2   & 31.2   \\
 & ASIS (\textit{vanilla}) & 43.0 & 46.3 & 50.6 & 39.2 \\
 &  ASIS & {\bf 44.7}  & {\bf 48.2} & {\bf 53.2} & {\bf 40.7} \\
 \hline
  \multirow{2}{*}{PN++}  
  & ASIS (\textit{vanilla}) & 49.6 & 53.4 & 62.7 & 45.8 \\
 & ASIS & {\bf 51.2}  & {\bf 55.1} & {\bf 63.6} & {\bf 47.5} \\
\hline
\end{tabular}
\end{center}
\caption{Instance segmentation results on S3DIS dataset.}
\label{tab:s3dis_ins_results}
\end{table}

\begin{table}[!hbt]
\small 
\begin{center}
\begin{tabular}{c|c|c|c|c}
\hline
\hline
    Backbone & Method    & mAcc    & mIoU & oAcc   \\
\hline
\hline 
\multicolumn{5}{c}{Test on Area 5} \\
\hline
\multirow{3}{*}{PN} 
 & PN (\textit{RePr}) & 52.1   &  43.4 & 83.5 \\
 & ASIS (\textit{vanilla}) & 52.9 & 44.7 & 83.7 \\
 & ASIS &  {\bf 55.7} & {\bf 46.4} & {\bf 84.5} \\
\hline 
\multirow{2}{*}{PN++} 
 & ASIS (\textit{vanilla}) & 58.3 & 50.8 & 86.7 \\
 & ASIS &  {\bf 60.9} & {\bf 53.4} & {\bf 86.9} \\
\hline 
\hline 

\multicolumn{5}{c}{Test on 6-fold CV} \\
\hline
\multirow{4}{*}{PN} 
 & PN~\cite{Qi_2017_CVPR} &  -  & 47.7 & 78.6 \\
 & PN (\textit{RePr}) &  60.3  & 48.9  & 80.3 \\
 & ASIS (\textit{vanilla}) & 60.7 & 49.5 & 80.4 \\
 & ASIS &  {\bf 62.3} & {\bf 51.1} & {\bf 81.7} \\
 \hline
\multirow{2}{*}{PN++} 
 & ASIS (\textit{vanilla}) & 69.0 & 58.2 & 85.9 \\
 & ASIS &  {\bf 70.1} & {\bf 59.3} & {\bf 86.2} \\
\hline
\end{tabular}
\end{center}
\caption{Semantic segmentation results on S3DIS dataset.}
\label{tab:s3dis_sem_results}
\end{table}

\subsection{S3DIS Results}

We conduct experiments on S3DIS dataset using PointNet and PointNet++ (single-scale grouping) as our backbone networks.
If no extra notes, our main analyses are based on PointNet.

\subsubsection{Baseline Method}
We report instance segmentation results of our baseline method in Table~\ref{tab:s3dis_ins_results}.
Based on PointNet backbone, our method achieves $46.3$ mWCov when evaluate by 6-fold cross validation, which shows an absolute 5.5-point improvement over the state-of-the-art method SGPN\footnote{We reproduced the results of SGPN using the code \href{https://github.com/laughtervv/SGPN}{at github}, published by the authors.}.
The superiority is  consistent across the four evaluation metrics.
Semantic segmentation results are shown in Table~\ref{tab:s3dis_sem_results}.
The mIoU of training without instance segmentation branch is $48.9$, which can be regarded as the result of pure backbone PointNet. 
Equipped with instance segmentation training, our semantic segmentation baseline result achieves $49.5$ mIoU, which is slightly better.
It indicates that the supervision of instance segmentation helps learn more general shared feature representation.
As for the training time, SGPN needs $16\sim17$ hours (excluding pre-training) to converge, while it only takes $4\sim5$ hours for our method to train from scratch, both on a single GPU.
More computation time comparisons can be referred to Table~\ref{tab:s3dis_efficiency}.
Our baseline method is demonstrated to be effective and efficient.

\subsubsection{ASIS}
\begin{figure}[!tb]
\includegraphics[width=0.48\textwidth]{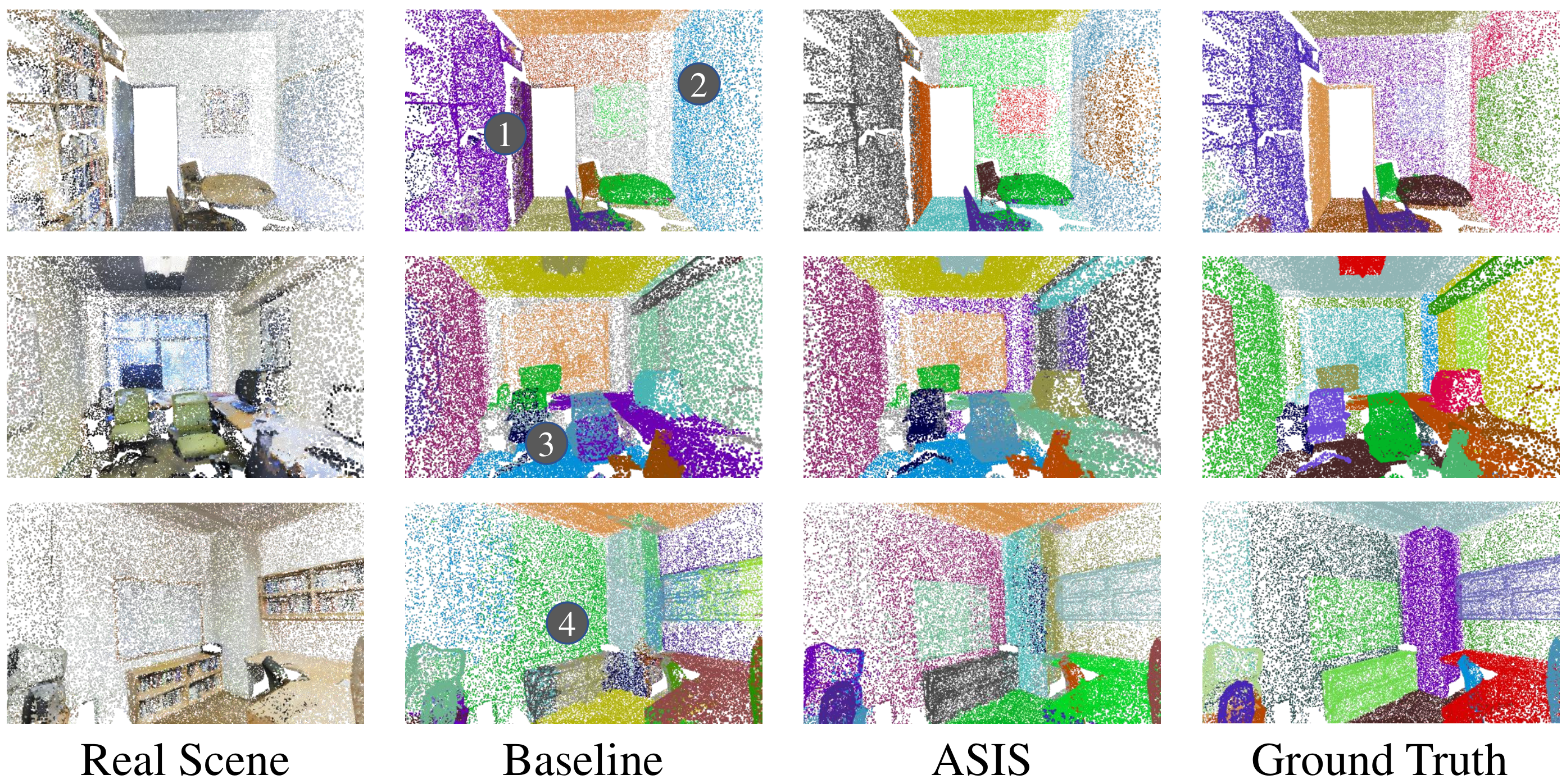}
\caption{Comparison of our baseline method and ASIS on instance segmentation. Different colors represent different instances.}
\label{fig:case_analysis_ins}
\end{figure}

\myparagraph{Semantic Segmentation.}
In Table ~\ref{tab:s3dis_ins_results},  we report the results of ASIS on instance segmentation task.
ASIS yields $48.2$ mWCov, which outperforms our baseline by 1.9-point.
In terms of another metric mean precision, a larger $2.6$-point gain is observed.
When evaluated on Area 5, the improvements are more significant: $2.7$ mWCov and $2.2$ mPrec.
Through visualizations in Figure~\ref{fig:case_analysis_ins}, our baseline method tends to group two nearby different class instances together into one instance (\eg, board \& wall). With ASIS, they are well distinguished as semantic awareness helps repel them in instance embedding space.
Per class performance changes are in accordance with our observations.
Shown in Table~\ref{tab:s3dis_perclass_results}, ASIS yields $5.0$ WCov and $2.4$ WCov gains on class ``board'' and class ``wall'' on instance segmentation. 

\begin{figure}[!tb]
\includegraphics[width=0.48\textwidth]{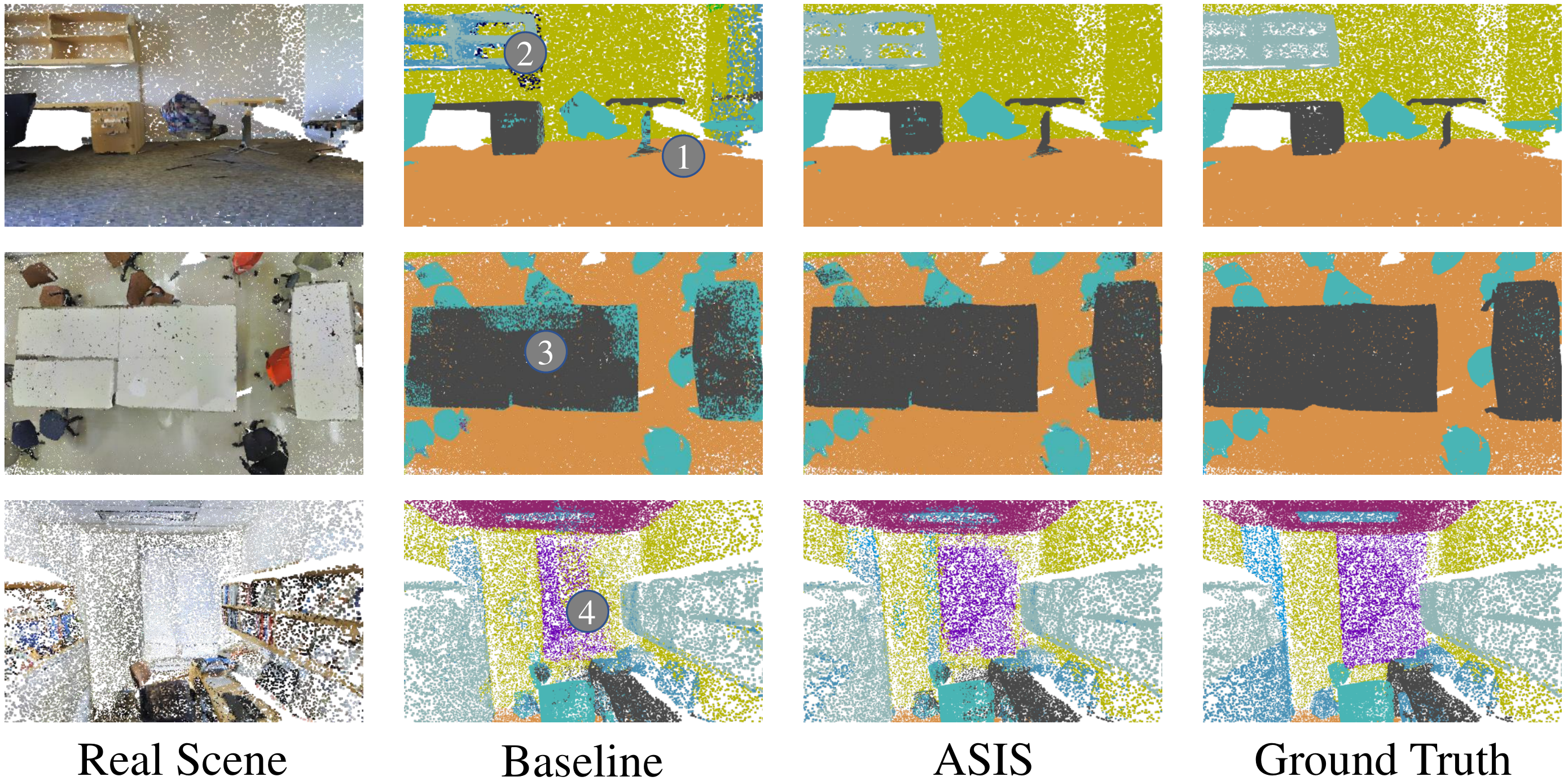}
\caption{Comparison of our baseline method and ASIS on semantic segmentation.}
\label{fig:case_analysis_sem}
\vspace{-0.2cm}
\end{figure}

\myparagraph{Instance Segmentation.}
Table~\ref{tab:s3dis_sem_results} reports the results of ASIS on semantic segmentation task.
ASIS improves the mIoU by 1.6-point.
We obseve more significant improvements of $2.8$ mAcc and $1.7$ mIoU when evaluating on Area 5.
In Figure~\ref{fig:case_analysis_sem} we show some comparison examples on semantic segmentation.
ASIS performs better on complicated categories (\eg, bookcase) and is aware of instance integrity (\eg, table, window) as instance fusion aggregates points belonging to the same instance to produce more accurate predictions.
Table~\ref{tab:s3dis_perclass_results} shows that ASIS outperforms the baseline by $3.5$ IoU and $2.2$ IoU on class ``table'' and class ``bookcase'', which are in line with our analysis.

\myparagraph{Stronger Backbone.}
Both the two tasks benefit largely from our novel method.
When adopt the stronger architecture PointNet++ as our backbone network, we observe consistent improvements: $2.1$ mWCov and $2.6$ mIoU gains on Area 5; $1.7$ mWCov and $1.1$ mIoU gains for 6-fold cross validation.
The results on PointNet++ indicate that our ASIS is a general framework and can be built upon different backbone networks.

\begin{table}[!t]
\small 
\begin{center}
\begin{tabular}{c|cc|c|c}
\hline
\hline
    Method & +IF & +SA &  mIoU & mWCov   \\
\hline
\hline
Baseline  & & & 49.5 & 46.3  \\
\hline
 & \checkmark & & 50.0 & 47.0 \\
 & & \checkmark & 49.8 & 47.4   \\
 & \checkmark & \checkmark & {\bf 51.1} & {\bf 48.2} \\
\hline
\end{tabular}
\end{center}
\vspace{-0.2cm}
\caption{Ablation study on the S3DIS dataset. IF refers to instance fusion; SA refers to semantic awareness.}
\label{tab:s3dis_ablation_results}
\end{table}

\begin{figure*}[!tb]
\includegraphics[width=0.98\textwidth]{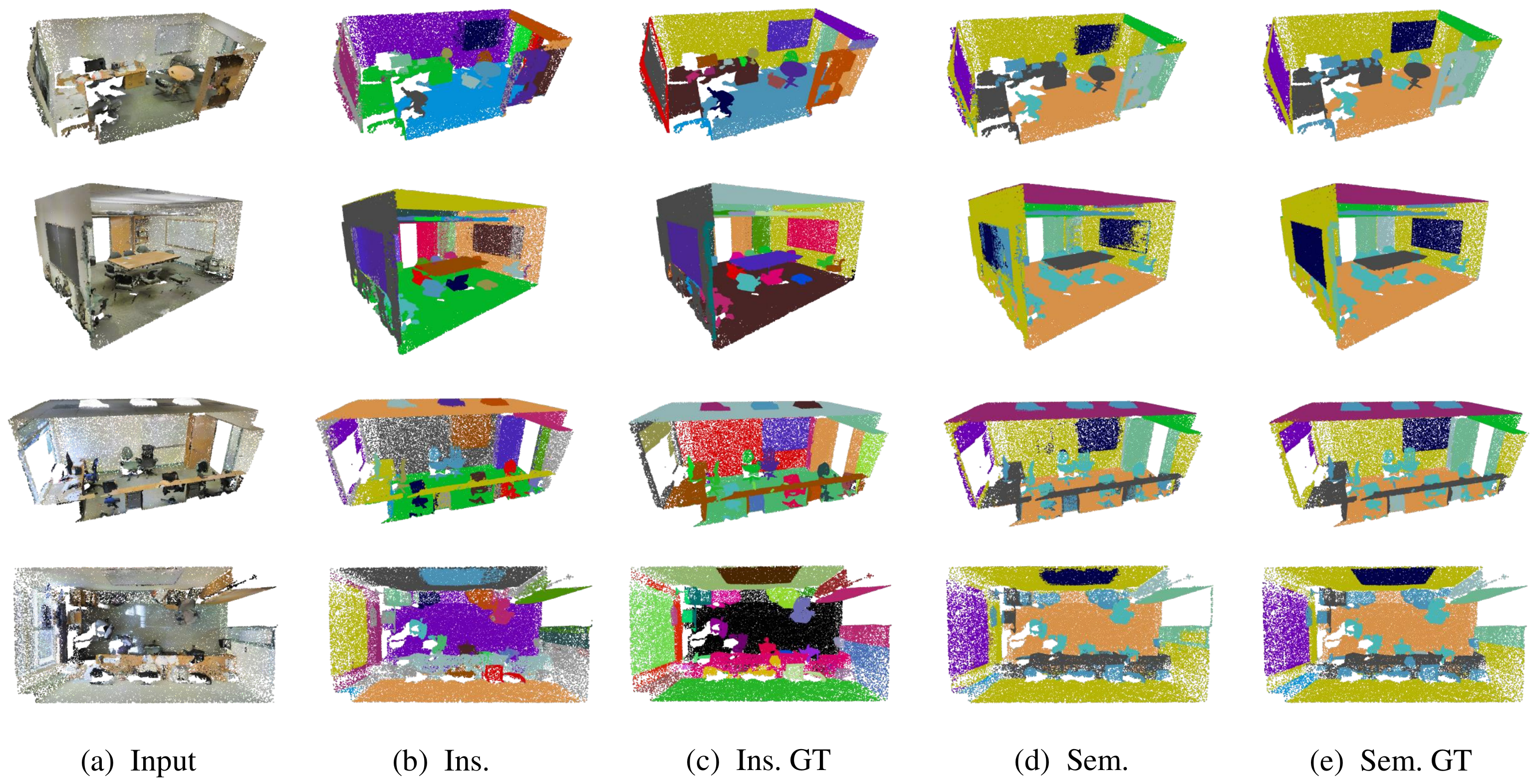}
\caption{Qualitative results of ASIS on the  S3DIS test fold.}
\label{fig:finalvis}
\vspace{-0.3cm}
\end{figure*}

\subsubsection{Analysis}
\label{subsubsec:ablation}
\myparagraph{Ablative Analysis.}
Equipped with only instance fusion for semantic segmentation, our method achieves $50.0$ mIoU and $47.0$ mWCov. Compared to the baseline, there is a 0.5-point gain on mIoU. 
Furthermore, better semantic predictions assign more correct class labels to instances, improving the instance segmentation performance.
When adopt semantic awareness alone, we achieve an improvement of $1.1$ mWCov (from $46.3$ to $47.4$). 
The improvement of one task also helps the other one, as better shared features are learned.  
Applying instance fusion and semantic awareness together, the performance boost is larger than using only one of them. 
On the basis of instance fusion, semantic awareness could bring additional $1.1$ mIoU and $1.2$ mWCov gains.
The semantic awareness strengthens the instance segmentation, as well as improving the semantic segmentation.
It is because that the improved instance embedding predictions could amplify the improvements brought by instance fusion, thus leading to a further $1.1$ mIoU gain.
The similar results can also be observed when add instance fusion on semantic awareness.
To conclude, the two components not only perform their own duty well, but also enlarge the function of the other one.

\begin{table*}[!t]
\begin{center}
\small 
\setlength{\tabcolsep}{3.8pt}
\begin{tabular}{c|c|ccccccccccccc}
\hline
        & mean   & ceiling  & floor & wall & beam & column & window & door & table & chair & sofa & bookcase & board & clutter \\
\hline
WCov & 
\begin{tabular}{@{}c@{}}46.3 \\ {\bf 48.2}\end{tabular} &
\begin{tabular}{@{}c@{}}79.2 \\ {\bf 80.1}\end{tabular} &
\begin{tabular}{@{}c@{}}{\bf 77.0} \\ 76.4\end{tabular} &
\begin{tabular}{@{}c@{}}63.7 \\ {\bf 66.1}\end{tabular} &
\begin{tabular}{@{}c@{}}47.6 \\ {\bf 53.4}\end{tabular} &
\begin{tabular}{@{}c@{}}6.6 \\ {\bf 9.2}\end{tabular} &
\begin{tabular}{@{}c@{}}55.6 \\ {\bf 58.8}\end{tabular} &
\begin{tabular}{@{}c@{}}47.5 \\ {\bf 49.8}\end{tabular} &
\begin{tabular}{@{}c@{}}50.5 \\ {\bf 50.6}\end{tabular} &
\begin{tabular}{@{}c@{}}57.3 \\ {\bf 59.4}\end{tabular} &
\begin{tabular}{@{}c@{}}{\bf 9.9} \\ {\bf 9.9}\end{tabular} &
\begin{tabular}{@{}c@{}}31.3 \\ {\bf 32.3}\end{tabular} &
\begin{tabular}{@{}c@{}}33.7 \\ {\bf 38.7}\end{tabular} &
\begin{tabular}{@{}c@{}}41.5 \\ {\bf 42.0}\end{tabular} 
\\
\hline
\hline
Sem IoU &  
\begin{tabular}{@{}c@{}}49.5 \\ {\bf 51.1}\end{tabular} &
\begin{tabular}{@{}c@{}}90.1 \\ {\bf 91.3}\end{tabular} &
\begin{tabular}{@{}c@{}}87.8 \\ {\bf 89.7}\end{tabular} &
\begin{tabular}{@{}c@{}}69.2 \\ {\bf 69.8}\end{tabular} &
\begin{tabular}{@{}c@{}}42.3 \\ {\bf 45.8}\end{tabular} &
\begin{tabular}{@{}c@{}}26.0 \\ {\bf 27.0}\end{tabular} &
\begin{tabular}{@{}c@{}}50.4 \\ {\bf 51.9}\end{tabular} &
\begin{tabular}{@{}c@{}}54.9 \\ {\bf 55.1}\end{tabular} &
\begin{tabular}{@{}c@{}}57.5 \\ {\bf 61.0}\end{tabular} &
\begin{tabular}{@{}c@{}}45.8 \\ {\bf 49.3}\end{tabular} &
\begin{tabular}{@{}c@{}}8.9 \\ {\bf 9.1}\end{tabular} &
\begin{tabular}{@{}c@{}}38.0 \\ {\bf 40.2}\end{tabular} &
\begin{tabular}{@{}c@{}}33.4 \\ {\bf 33.5}\end{tabular} &
\begin{tabular}{@{}c@{}}39.2 \\ {\bf 40.7}\end{tabular} 
\\
\hline
\end{tabular}
\end{center}
\vspace{-0.3cm}
\caption{Per class results on the S3DIS dataset.}
\vspace{-0.1cm}
\label{tab:s3dis_perclass_results}
\end{table*}

\begin{table}[!ht]
\small 
\begin{center}
\setlength{\tabcolsep}{3.8pt}
\begin{tabular}{c|c|cc|c}
\hline
\hline
 \multirow{2}{*}{Method}      &   \multicolumn{3}{c|}{Inference Time (ms)} &  \multirow{2}{*}{mWCov}\\
\cline{2-4}
 & Overall & Network & Grouping & \\
\hline
SGPN       & 726  & 18  & 708  & 35.5   \\
ASIS (\textit{vanilla})  & 212 & 11 & 201 & 41.4 \\
ASIS      & 205 & 20 & 185 & 43.6 \\
\hline
ASIS (\textit{vanilla}.PN++)  &  150 & 35 & 115 & 45.7 \\
ASIS (PN++)  & 179 & 54 & 125 &  47.8 \\

\hline
\end{tabular}
\end{center}
\vspace{-0.3cm}
\caption{Comparisons of computation speed and performance. Inference time is estimated and averaged on Area 5, which is the time to process a point cloud with size $4096 \times 9$. The instance segmentation results on Area 5 are reported. }
\label{tab:s3dis_efficiency}
\vspace{-0.2cm}
\end{table}

\begin{figure}[htbp]
\centering
\subfigure[]{
\includegraphics[width=0.22\textwidth]{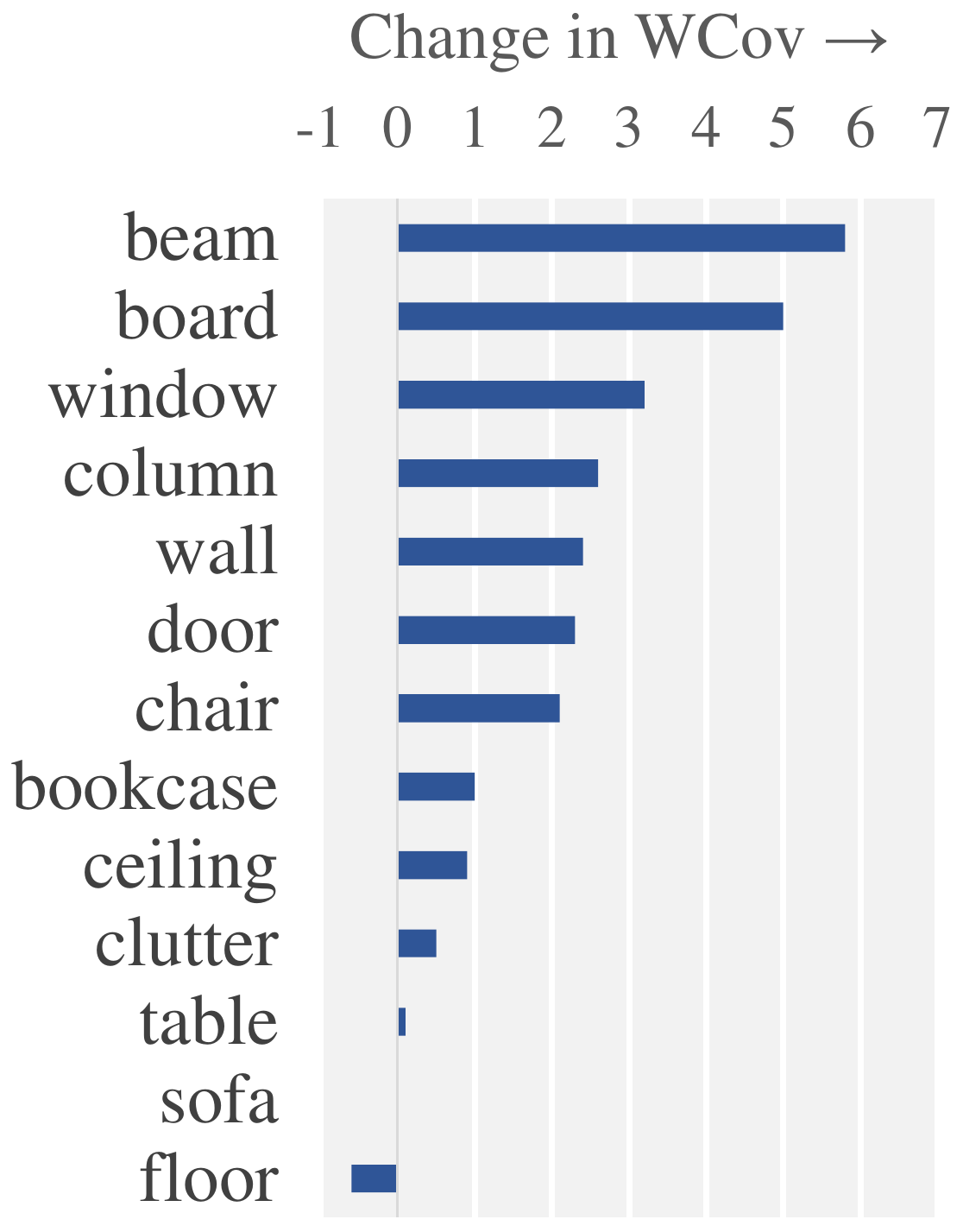}
\label{fig:percate_wcov}
}
\subfigure[]{
\includegraphics[width=0.22\textwidth]{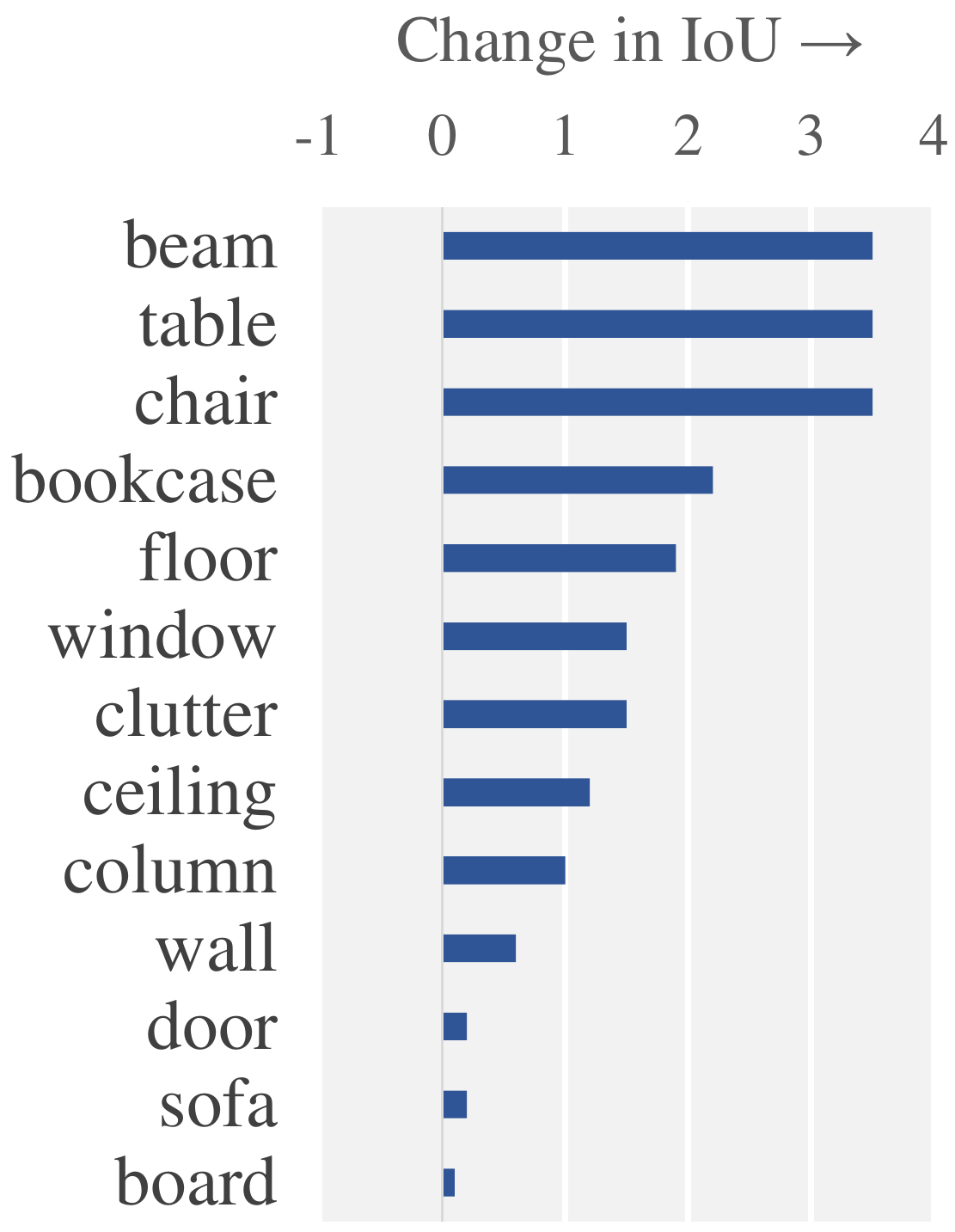}
\label{fig:percate_iou}
}
\vspace{-0.1cm}
\caption{Per class performance changes. (a) Changes of instance segmentation performance compared to our baseline method. (b) Changes of semantic segmentation performance compared to our baseline method.}
\label{fig:percate}
\vspace{-0.2cm}
\end{figure}

\myparagraph{Category-based Analysis.}
We show how the performance of each category changes in Figure~\ref{fig:percate}.
Interestingly the categories being helped by ASIS module are different for instance segmentation and semantic segmentation.
On instance segmentation, our ASIS module largely helps the categories in which  instances often surround with instances of other classes (\eg, beam, board and window). 
For example, board is hung on the wall.
The board is easily being ignored during instance segmentation, as the body of the board has similar color and shape with the wall.
Our semantic awareness in ASIS module shows great superiority on these cases: $5.0$ WCov and $2.4$ WCov improvements on class ``board'' and class ``wall''.  
Some visualization examples of the comparison are illustrated in Figure~\ref{fig:case_analysis_ins}.
On semantic segmentation, ASIS module significantly boosts the performance of the categories in which instances have complicated shapes (\eg, table, chair and bookcase), because they benefit much from instance fusion.

\vspace{-0.2cm}
\subsubsection{Qualitative Results}
Figure~\ref{fig:finalvis} shows some visualization examples of ASIS.
For instance segmentation, different colors represent different instances, while the color itself does not mean anything.
Either same class instances or different class instances are distinguished properly.
For example, the points of the tables and the surrounding chairs are grouped into distinct instances. 
As for semantic segmentation, specific color refers to particular class (\eg, yellow for wall, purple for window).
We also show some failure cases in Figure~\ref{fig:finalvis}.
In the scenes of the second and third row, two nearby chairs are mistakenly segmented together as a single instance.
Though our method does not draw the point embeddings of the same class instances close, we yet do not contribute on better distinguishing this kind of cases.  
We leave it to future works to explore better solutions.

\vspace{-0.2cm}
\subsubsection{Computation Time}
In Table~\ref{tab:s3dis_efficiency}, we report computation time  measured on a single Tesla P40 GPU. 
The inference procedure can be divided in to two steps: the network inference, and point grouping which groups points into individual instances.
For SGPN, the grouping step refers to their GroupMerge algorithm. 
In our ASIS, it is the mean-shift clustering.
We achieve comparable speed with SGPN on network inference, while our grouping step is much faster.
Overall, it takes $205$ms for ASIS to process an input point cloud with size $4096\times9$ and output the final labels, which is $3.5\times$ faster than SGPN.

\subsection{ShapeNet Results}
\begin{figure}[!tb]
\includegraphics[width=0.48\textwidth]{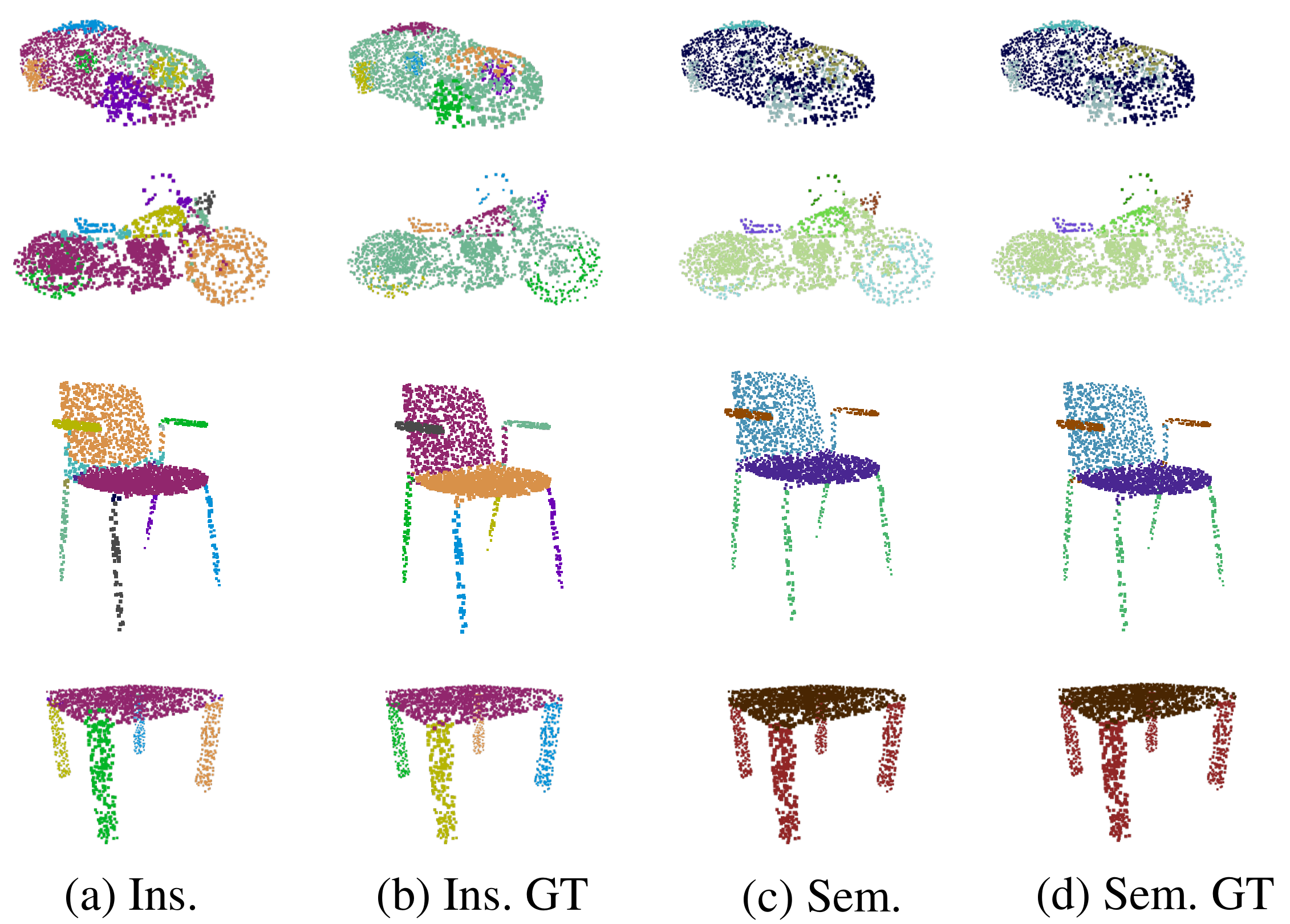}
\caption{Qualitative results of ASIS on ShapeNet test split. (a) Instance segmentation results of ASIS. (b) Generated ground truth for instance segmentation. (c) Semantic segmentation results of ASIS. (d) Semantic segmentation ground truth.}
\label{fig:shapenet}
\vspace{-0.2cm}
\end{figure}

We conduct experiments on ShapeNet dataset using instance segmentation annotations generated by~\cite{sgpn}, which are not ``real'' ground truths. 
Following~\cite{sgpn}, only qualitative results of part instance segmentation are provided.
As shown in Figure~\ref{fig:shapenet}, tires of the car and legs of the chair are well grouped into individual instances.
Semantic segmentation results are reported in Table~\ref{tab:shapenet_sem_results}.
Using PointNet as backbone, we achieve a 0.6-point improvement.
Based on PointNet++, ASIS outperforms the baseline by 0.7 mIoU. 
These results demonstrate that our method is also beneficial for part segmentation problem.

\begin{table}[!ht]
\small
\begin{center}
\begin{tabular}{c|c}
\hline 
\hline
 Method       & mIoU \\
\hline
PointNet~\cite{Qi_2017_CVPR} &  83.7 \\
PointNet (\textit{RePr}) &  83.4 \\
PointNet++~\cite{qi2017pointnet++}* &  84.3 \\
\hline
ASIS (PN) &   84.0 \\
ASIS (PN++) &  85.0 \\
\hline
\end{tabular}
\end{center}
\vspace{-0.3cm}
\caption{Semantic segmentation results on ShapeNet datasets. \textit{RePr} is our reproduced PointNet. PointNet++* denotes the PointNet++ trained by us without extra normal information.}
\label{tab:shapenet_sem_results}
\end{table}

\section{Conclusion}
    In this paper, a novel segmentation framework, namely ASIS, is proposed for associating  instance segmentation and semantic segmentation on point clouds.
    The relationships between the two tasks are explicitly explored and directly guide our method design.
    Our experiments on S3DIS dataset and ShapeNet part dataset demonstrate the effectiveness and efficiency of ASIS.    
    We expect wide application of the proposed method in 3D instance segmentation and 3D semantic segmentation, as well as hoping the novel design provides insights to future works on segmentation tasks, \eg, panoptic segmentation, and beyond.

{\small
\bibliographystyle{ieee}
\bibliography{egbib}
}

\end{document}